# Human Action Recognition System using Good Features and Multilayer Perceptron Network

Jonti Talukdar, Bhavana Mehta

*Abstract*—Human action recognition involves the characterization of human actions through the automated analysis of video data and is integral in the development of smart computer vision systems. However, several challenges like dynamic backgrounds, camera stabilization, complex actions, occlusions etc. make action recognition in a real time and robust fashion difficult. Several complex approaches exist but are computationally intensive. This paper presents a novel approach of using a combination of good features along with iterative optical flow algorithm to compute feature vectors which are classified using a multilayer perceptron (MLP) network. The use of multiple features for motion descriptors enhances the quality of tracking. Resilient backpropagation algorithm is used for training the feedforward neural network reducing the learning time. The overall system accuracy is improved by optimizing the various parameters of the multilayer perceptron network.

*Index Terms*—Action Recognition, Artificial neural network, Image motion analysis, Multilayer perceptron.

## I. INTRODUCTION

HUMAN action recognition (HAR) is the novel problem of characterizing human actions through the automated analysis of video sequences. This can be accomplished by the use of a variety of image processing and machine learning algorithms. HAR is based on the concept of using the output of a standard video camera to study, localize and analyze the actions being performed by the human subject. Within the past decade, a huge amount of interest has been generated in this field due to its challenging nature and its wide variety of applications. Recognizing human actions is an integral step in improving the domain of human-computer interaction and poses significant challenges which need to be overcome.

Several parameters like dynamic backgrounds, moving cameras, multiple subjects, scattered environments, variable aspect ratios, occlusions as well as variable ambient lighting conditions pose hurdles in accurate action recognition [1]. These challenges make it difficult to achieve a system which uses limited computing resources and still performs action recognition in a robust fashion.

Several complex approaches to solve these problems have been highlighted in recent times, however, all these methods are heavily reliant on the use of local motion descriptors like space-time interest points (STIPs) [2], pose estimation [3], spatio-temporal gradients etc. for learning, classifying and recognizing human actions. These methods however, are computationally intensive and complex, requiring the use of a large amount of training data for classification and thus add to the challenge of achieving action recognition in real time [4]. Several new and promising approaches for action recognition have also surfaced in recent times, due to the introduction of newer and advanced technologies for video and motion capture. The development of complex sensor networks and data acquisition devices including IR cameras, motion sensors, stereo cameras etc. can also contribute to the development of HAR systems, but necessitate the use of specialized equipment and hence are not feasible for large scale deployment. Thus, to facilitate wide scale adoption and deployment of HAR systems, it is crucial to overcome the above mentioned setbacks and design a system with significant reduction in computational complexity as well as the use of cheap and inexpensive components.

HAR finds potential applications in diverse domains including video surveillance, sign language recognition, sixth sense technology, search and rescue, video indexing, psychometric testing, smart cities etc. Thus creating a HAR system which is both computationally efficient as well as sufficiently accurate and robust will lead to its successful adoption in various fields like security, healthcare, sports etc.

In this paper, we propose a simple yet efficient HAR system which can be easily implemented on a single board computer and hence deployed cheaply anywhere, anytime. The proposed HAR system uses a combination of the iterative optical flow algorithm along with 'good features' [5] as the primary motion descriptors. The use of this approach ensures that multiple features are selected for motion description which ultimately enhances the quality of tracking of the subject of interest. Tracking of good features in dynamic video frames is also easier and computationally inexpensive as opposed to performing various transformations of entire frames in spatial and temporal domains [5]. Thus the above approach ensures that the system operates in real time using limited computing resources. The system has been designed to distinguish between four actions which are walking, running, boxing and clapping. This however, can be extended to accommodate a more diverse set of action classes depending on the comprehensiveness of the dataset being used or created as per user requirements. The

Jonti Talukdar is with Department of Electronics and Communication Engineering, Institute of Technology, Nirma University, Ahmedabad, Gujarat, India. (e-mail: 14bec057@nirmauni.ac.in).

Bhavana Mehta is with Department of Electronics and Communication Engineering, Institute of Technology, Nirma University, Ahmedabad, Gujarat, India. (e-mail: 14bec028@nirmauni.ac.in).

KTH action dataset [1] is used for training and classification of the model. A multilayer perceptron (MLP) network, which is based on feedforward artificial neural network is used for supervised learning as well as classification. The proposed combination of features offers the advantage of being computationally simple as compared to earlier methods, yet being robust in action detection when used in conjunction with the MLP based feedforward neural network classifier. The overall system accuracy is increased by optimizing the number of feature vectors per sample, number of hidden nodes in MLP as well as total training samples.

The remainder of the paper discusses the various aspects of the system and is organized as follows: Section II details the various related work done in the domain of action recognition, Section III highlights the salient features of the proposed HAR system including the final HAR algorithm which uses good features in conjunction with the iterative optical flow algorithm, Section IV presents a summary of simulation results and discusses about system optimization. Finally, Section V concludes the paper.

## II. RELATED WORK

In this section, we briefly discuss about the various approaches adopted in recent times to solve the problem of action recognition. The overall objective of a HAR system is the automated analysis and interpretation of ongoing events and their context from video data [6]. Thus, the overall performance of the HAR system narrows down to the proper selection of feature vectors for action classification. Poppe [7] has divided image feature representations for HAR into two parts, global representations and local representations. The former is faster to compute and encodes the visual information of the whole image frame as feature vectors while the latter is more robust but computationally intensive and identifies specific patches of local activity around temporal interest points. Several successful methods which are based on the local approach for selection and evaluation of action specific features have been developed in recent times and are discussed below.

The use of silhouettes for tracking motion between multiple frames was pioneered by Bobick and Davis [8]. They used aggregate differences in video frames due to the dynamic nature of video data and generated a scalar field consisting of pixels whose intensities were a function of recent motion. This is called motion history image (MHI) and forms an entire stack of silhouettes within the space-time volume. Template matching is then performed for classification by evaluating Hu moments.

Schüldt et al. [1] and Laptev et al. [9] described the use of representing motions in the form of local space-time interest points. A scale-space representation of the image sequence is first obtained by using a Gaussian convolution kernel. Local features are then detected by computing the second moment matrix of the scale space representation within a Gaussian neighborhood of each point respectively. The local feature points obtained are classified using support vector machine (SVM). The results obtained on the test set are robust to variations in environment as well as noise.

An unsupervised learning approach was highlighted by Niebles et al. [10] in which Space-time interest points [2] are evaluated for local regions within the overall image sequence. The results of these points are then combined together and clustered to create a codebook for each class of action. Recognition is done by performing feature extraction on the input and then performing latent topic discovery through two models, probabilistic latent semantic analysis [11] and latent dirichlet allocation [12].

Recently, other classification techniques like boosting which combine several weak classifiers to form a stronger one have been proposed by Zhang et al. [13]. The use of convolutional neural network (CNN) for localization and tracking of key points on image sequences was also highlighted by Fan et al. in [14]. A CNN based hybrid feature selection method in the form of motion capture through depth cameras was also proposed by Ijjina and Mohan [15].

Since all these approaches used local descriptors, the results though highly accurate, could not be achieved in real time leading to significant delay. Hence, we have used a global approach which uses the strongest features in the video sequence in conjunction with the iterative optical flow algorithm to track and classify human actions. Details about the proposed method are discussed in Section III.

## III. PROPOSED METHOD

We propose a novel HAR system which uses both interest point based features in the form of good features to track as well as motion based features in the form of optical flow for feature detection within image sequences. Fig. 1 describes the algorithmic steps (system flow) for the proposed HAR system.

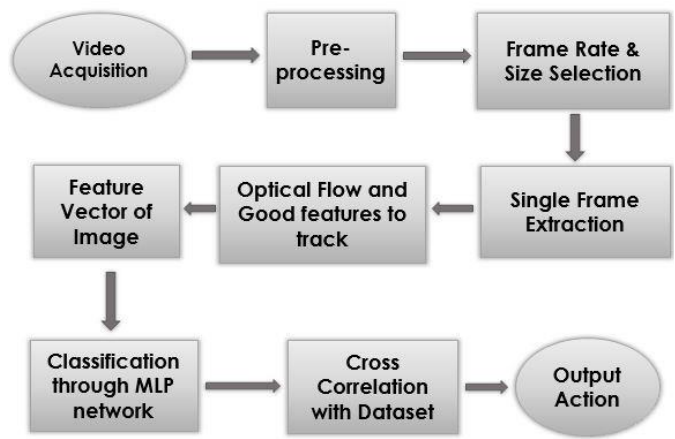

Fig. 1. Proposed HAR algorithm.

The overall algorithm can be further subdivided into the three major stages which include (i) Preprocessing, (ii) Feature extraction and (iii) Classification. These stages along with their salient features are discussed below in detail.

### A. Preprocessing

The video data acquired from the camera in crude form needs to be preprocessed in order to make it useful irrespective of its source. Preprocessing also removes unnecessary parts of the image sequences thus reducing the data size to work on, thus also decreasing the processing time and improving overall system speed. It also removes unwanted noise from images thus making the classifier more accurate. Fig. 2 shows the principal steps of the preprocessing stage, which are described below in more detail.

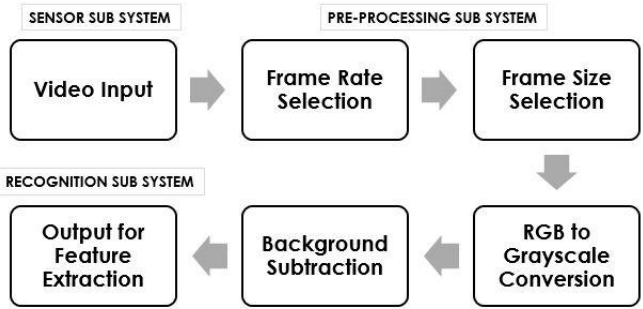

Fig. 2. Architectural description of the preprocessing stages.

*1) Frame Rate and Size selection*: The foremost operation performed for recognizing actions from the video sequence is to set the frame size and frame rate of that particular test or training video. This is done to maintain dimensional uniformity of the feature vector file generated for both the training as well as the testing videos respectively. The input video is downsampled and resized to a spatial resolution of 160*120 pixels because the videos present in the KTH dataset used for training the HAR system [1] have the same spatial resolution. Decreasing the video size also serves the added purpose of increasing the processing speed because of the decreased number of pixels on which feature selection is performed.

*2) RGB to Grayscale conversion:* Grayscaling is performed on the downsampled video by evaluating the average contributions from each of the three channel values (R, G and B) respectively. Grayscaling further reduces the size and feature vectors of the video by a considerable amount. This is because all the three channels of the RGB colorspace represent 24 bits where 8 bits corresponds to each of the primary color. This however, after grayscaling operation, is reduced to only 8 bits for the entire colorspace. Thus, the size reduction leads to better memory management and also improves the overall system performance.

*3) Background Subtraction:* In context of HAR, background subtraction is used as a form of image segmentation technique to localize areas of motion within the frame. Building upon the rationale of detecting moving objects by subtracting consecutive frames, i.e. the difference between current frame and the reference frame, a Gaussian mixture model based background subtraction technique as proposed in [16] is used. Since each background pixel is modelled by a mixture of $k$ different Gaussian distributions, the probability that a pixel has a value $x$ at time $N$ is given as follows [16]:

$$\Pr(x) = \sum_{k=1}^{K} w_k N(x; \mu_k, \varepsilon_k) \quad (1)$$

where $N(x; \mu_k, \varepsilon_k)$ denotes the normal distribution of the $k^{th}$ component with mean $\mu_k$ and variance $\varepsilon_k$:

$$N(x; \mu_k, \varepsilon_k) = \frac{1}{2\pi^{\frac{d}{2}}|\Sigma_k|^{\frac{1}{2}}} e^{-\frac{1}{2}(x-\mu)(x-\varepsilon)} \quad (2)$$

The *fitness value* given by $w_k/\varepsilon_k$ gives the measure of formation of static clusters within the frame. Thus, background subtraction is evaluated by measuring each new pixel against the reference frame on the basis of its fitness value [16]. The advantage of using a Gaussian mixture model is that it is computationally inexpensive and we can clearly estimate as well as distinguish the probability that a patch belongs to a particular frame or region of interest.

*B. Feature Extraction*

The overall recognition rate of the HAR system directly depends on the unique features that are computed and tracked per frame. Thus, higher the quality of the features that are tracked, higher is the accuracy of the system. As a result, 'good features' [5], which is a modification of the corner detector algorithm [17] is evaluated on the preprocessed downsampled image sequences from the previous steps. The feature selection model as proposed by [5] is based on evaluating and monitoring interest point features whose selection maximizes the quality of tracking of the desired subject. The quality of tracking is not only measured by the degree of "cornerness" of the feature but also by differentiating between good features and bad features based on a dissimilarity index, which is evaluated as the rms residue of the feature in question between two consequent frames. An image sequence $I(x, y, t)$ under motion can be described as [5]:

$$I(x, y, t + \tau) = I(x - \xi(x, y, t, \tau), y - \eta(x, y, t, \tau)) \quad (3)$$

where the vector $\boldsymbol{\delta} = (\xi, \eta)$ is called the displacement vector at point $\boldsymbol{X} = (x, y)$. Image motion between two frames is better represented through *affine motion* given as [5]:

$$\delta = D\boldsymbol{X} + \boldsymbol{d} \quad (4)$$

where $D$ is the deformation matrix and $\boldsymbol{d}$ is the linear translation. Thus, if a point on an image frame $I$ translates to another point on the image frame $J$, the key task is to minimize the dissimilarity by finding the minimum value of $D$ and $\boldsymbol{d}$ respectively. Upon linearizing the dissimilarity and computing the difference between the two frames [5], the error vector is found to be directly dependent on a matrix, $Z$, where:

$$\boldsymbol{e} = Z\boldsymbol{d} \quad (5)$$

If $\lambda_1$ and $\lambda_2$ are the two eigenvalues of $Z$, then a good feature is selected only if:

$$min(\lambda_1, \lambda_2) > \lambda \quad (6)$$

where $\lambda$ is a threshold value [5]. Good features thus help in computing and tracking the $N$ strongest features in a dynamic image sequence. Using motion based features in conjunction with interest point features increases the accuracy of the system by many folds. Thus, the Lukas-Kanade iterative tracking algorithm [18] is applied to the good features extracted. The feature vector $\boldsymbol{F}(x, y, t)$ thus obtained can be represented as

follows [19]:

$$\boldsymbol{F}(x,y,t) = [x, y, t, I_t, u, v, u_t, v_t, D_{iv}V_{or}, G_{ten}, S_{ten}]^T \quad (7)$$

where $I(x, y, t)$ is the acquired image sequence, $\boldsymbol{u}(x, y, t)$ is the corresponding optical-flow vector, $I_t$ is the 1st order partial derivative of $I(x, y, t)$ with respect to $t$, $D_{iv}$ is the spatial divergence of the vector field, $V_{or}$ is the measure of local spin or vorticity of the flow fields, and $G_{ten}$ and $S_{ten}$ are invariant tensors [19] given as follows:

$$D_{iv}(x,y,t) = \frac{\partial u(x,y,t)}{\partial x} + \frac{\partial v(x,y,t)}{\partial y} \quad (8)$$

$$V_{or}(x,y,t) = \frac{\partial v(x,y,t)}{\partial x} - \frac{\partial u(x,y,t)}{\partial y} \quad (9)$$

$$G_{sten}(x,y,t) = \frac{1}{2}\left(tr(\nabla u(x,y,t)^2)\right) - tr(\nabla v(x,y,t)^2)) \quad (10)$$

$$S_{ten}(x,y,t) = \frac{1}{2}\left(tr(S(x,y,t)^2)\right) - tr(S(x,y,t)^2)) \quad (11)$$

The algorithm calculates the flow vector for each of the selected feature in a pyramidal fashion, i.e. motions are tracked according to their scale in an adaptive fashion. The tracking algorithm which works for small pixel movements fails for large scale motions. In such cases, the smaller motions are smoothed out and the larger ones are tracked iteratively. This algorithm works even in cases where the next feature is occluded, because the flow vector is evaluated using not just one feature but the neighboring pixels as well. This makes the system robust. It also allows us to compute the flow vector for every third frame, thus reducing the processing time as well.

*C. Classification*

The extracted feature vectors are passed through a feedforward artificial neural network for classification. The classifier, based on its characteristics, maps the given feature vector into each of the four possible action classes. The Multilayer perceptron (MLP), which is used as the primary classifier, is composed of multiple nodal layers which are interconnected in a directed graph fashion. MLP and other forms of neural network models are used in situations where traditional algorithmic computations for feature analysis and classification are too complex and the systems are thus trained rather than programmed. The MLP architecture is such, that every individual node or neuron consists of a sigmoid activation function [21]:

$$f(x) = \beta * (1 - e^{-\alpha x})/(1 + e^{-\alpha x}) \quad (11)$$

Thus the input feature vector, when passed through the multiple layers downstream, undergoes nonlinear transformation and becomes linearly separable [21]. For an arbitrary output $x_j$ of a layer $\eta$, the output for $\eta + 1$ layer is the sum of the individual weights of each neuron and a bias function [20]:

$$y_i = \sum_j \left(w_{i,j}^{n+1} * x_j\right) + w_{i,j}^{n+1} + f(u_i) \quad (12)$$

where $w_{i,j}^{n+1}$ is the weight of the individual neuron and $f(u_i)$ is the bias function. The resilient propagation (R-PROP) adaptive backpropagation algorithm [20] is used for training the MLP. As the model is trained, the weights of individual neurons are adapted locally based on their influence on an arbitrary error function. An update value $\Delta_{ij}$ for each $w_{ij}$ is evaluated based on the sign of partial derivatives of the error function in each dimension of the weight-space. This direct adaptation of the weight updates reduces the learning steps significantly, is very efficient to compute in terms of storage and time and is also robust against the choice of starting values [20], thus improving the overall system performance. The algorithm for the overall system is shown in Fig. 3.

| **Algorithm:** Algorithm for HAR |
|---|
| **Input:** Video stream from static camera |
| **Output:** Recognized action class |
| 1: Frame rate & size initialization = 160*120p. |
| 2: RGB to Grayscale conversion. |
| 3: **for each** Pr(x) at a given time $N$ **do** |
| 4: Evaluate fitness value $w_k/\varepsilon_k$. |
| 5: Subtract current frame and previous frame. |
| 6: **end for** |
| 7: **for each** Image sequence $I(x,y,t)$ **do** |
| 8: Evaluate deformation $D$ and linear translation $\boldsymbol{d}$. |
| 9: Initialize tracking parameter $\lambda$. |
| 10: **if** $min(\lambda_1, \lambda_2) > \lambda$ **then** |
| 11: Select feature for tracking. |
| 12: Evaluate feature vector $\boldsymbol{F}(x,y,t)$. |
| 13: **end if** |
| 14: **end for** |
| 15: Initialize number of feature vectors per frame, training samples, and number of hidden nodes in MLP. |
| 16: Evaluate individual neuron weights $w_{ij}$ during training stages by passing training video data. |
| 17: Pass feature vector $\boldsymbol{F}(x,y,t)$ to trained model for classification. |
| 18: **return** recognized action class. |

Fig. 3. Overall HAR pseudocode.

IV. SIMULATION AND RESULTS

In this section we evaluate the overall performance of our proposed HAR system. The KTH dataset, which has been used for training, consists of a repository of 2,391 videos of six different action classes performed by 25 different subjects [1]. Four action classes: walking, running, boxing and clapping have been used for training the MLP classifier. Recognition results have been evaluated for both real time videos as well as test videos from the dataset. A comparative analysis of the various parameters of the MLP classifier is done to ensure an optimized system easily implementable on a SBC.

*A. Feature Extraction Results*

As discussed in Section III, the two major features which are used in our proposed system include "good features" and the optical flow vectors. Simulation results for both these feature extraction techniques are shown below. Fig. 4 shows the comparative results obtained during the process of localizing and tracking good features in the input video frame for both clapping as well as walking. Good features, represented by blue dots, are tracked on the acquired input video and the number of good features is predetermined by the user, depending on the required accuracy. The feature size, which also includes the number of good features is an important factor when considering the overall system accuracy and is discussed in more detail below.

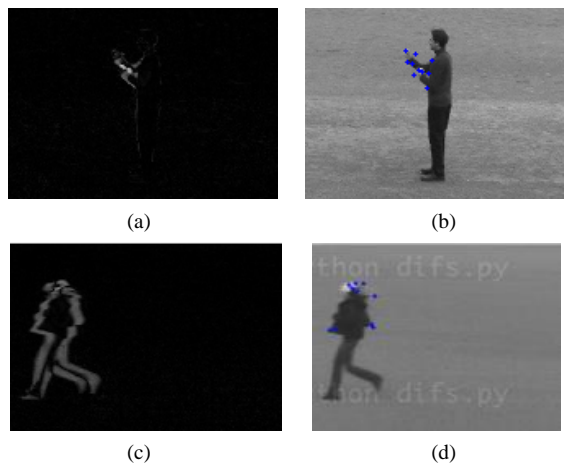

Fig. 4. (a) and (c) show the background subtraction results on clapping and walking where the grey silhouette represents dynamic characteristics of the frame. (b) and (d) show the good features tracked for the same action set where the blue dots indicate the good features.

The background subtraction results show that only areas in the neighborhood of the tracked features are dynamic in nature. Fig. 5 shows the visualization of motion flow within the input video, where the green arrows represent the flow vectors. It can be observed that the iterative optical flow algorithm when applied only to the good features tracked provide significant motion description of the movement of the subject and can be conclusively used for training and classification.

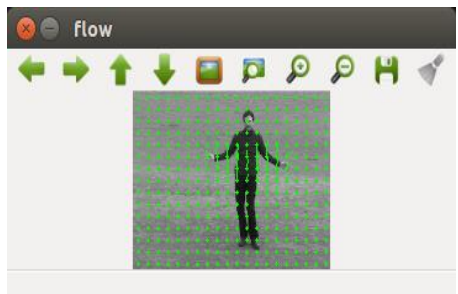

Fig. 5. Green arrows showing the motion flow vectors.

### B. Training and Classification Results

The recognition rate of the overall system is directly dependent on the size of the feature vector per frame, the total number of training samples used in the MLP classifier as well as the number of hidden layers present in the MLP network. However, increasing each and every parameter to the highest possible resolution will lead to a very slow system which is heavily dependent on processing resources. Thus it is essential to find a tradeoff between speed and accuracy to ensure that the overall purpose of the HAR system is achieved with respect to both robustness and real time results.

Fig. 6 shows the final output of the HAR system for the boxing action class in real time. The subject performs the action in front of the static camera. The feature vectors are evaluated for the real time video input through and then passed through an already trained MLP network. The predicted action class is displayed in the terminal with an update time of 1 second. The system works well in case of an uncluttered background with a frame rate of 25fps.

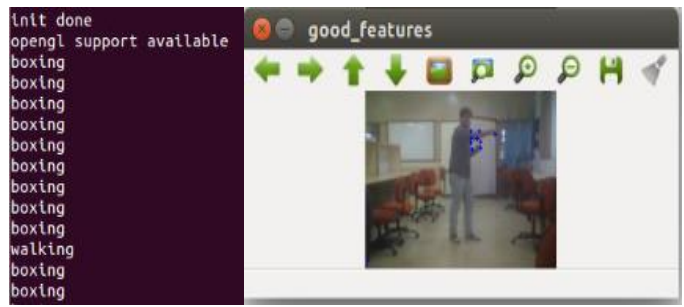

Fig. 6. Real time implementation of the HAR system for the boxing action class.

In order to decide on the optimum performance parameters of the system, it is essential to analyze the tradeoff between the different sizes of feature vector used for classification with their corresponding recognition rates respectively. Table I shows the recognition rates for the four action classes on a MLP consisting of 200 hidden nodes and a training set of 299 samples. It can be readily concluded from the table, that for any given action class, as the feature vector size increases, the recognition rate also increases. However, this relationship is nonlinear and there exists a point where increasing the feature size no longer improves the overall system performance. At this point, the lag or delay involved in processing the feature vectors outweighs the benefit in improved accuracy and the overall system performance reduces.

TABLE I
RECOGNITION RATE OF ACTION CLASSES

| Action Class | Recognition Rate | | |
|---|---|---|---|
| | Feature size 14 | Feature size 10 | Feature size 8 |
| Boxing | 95.2 | 93.2 | 89.8 |
| Clapping | 93.4 | 92 | 90.6 |
| Running | 95.2 | 94.3 | 89.4 |
| Walking | 94.6 | 93 | 88.4 |

This tradeoff point can be easily observed in Fig. 7, which shows the variation in system accuracy (recognition rate) with various feature vector sizes for all the four action classes. If we consider a base feature vector size of 10, then the increase in feature vector size by 40% increases the average accuracy of the HAR system by only 1.3%, whereas decreasing feature size by even 20% reduces the average accuracy by more than 3.5%. Thus for real time applications, a feature vector size of 10 is ideal. Similarly, keeping the other classifier parameters same, reducing the number of hidden nodes in the MLP network by 40% reduced the average accuracy of the system decreased from 92% to 51%. Also, an accuracy of more than 98% is obtained for the clapping action class by increasing the number of training samples to 400. However, this has the added disadvantage of increasing the training time.

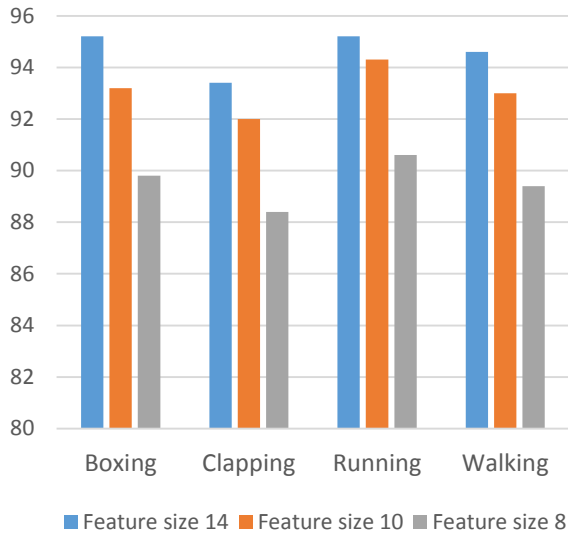

Fig. 7. Recognition rate for various action classes and feature sizes.

To judge the overall accuracy of the classification model, a confusion matrix for all the four action classes is drafted and shown in Table II. 120 test videos from each action class were tested and passed through the trained neural network for classification.

TABLE II
CONFUSION MATRIX FOR ALL ACTION CLASSES

|  | Boxing | Clapping | Running | Walking |
|---|---|---|---|---|
| Boxing | **112** | 7 | 0 | 1 |
| Clapping | 9 | **110** | 0 | 1 |
| Running | 1 | 0 | **113** | 6 |
| Walking | 2 | 0 | 7 | **111** |

It can be observed that an overall system accuracy of more than 92% is obtained with running and boxing action classes having higher recognition rates. The above analysis is essential in defining the design features of any HAR system. The final set of system parameters which can easily be implemented on a SBC consists of 200 hidden nodes for the MLP, a feature size of 10 and a training sample of 300 videos.

## V. CONCLUSION

Real time human action recognition implemented on a low cost yet efficient and accurate system (SBC) will find immense applications in the real world like smart video surveillance, sign language recognition, and unusual activity recognition. Further development of the proposed system will act as a technological incubator for applications in search and rescue situations, accessing remote areas, terrorist prone zones, disaster hit zones etc. In this work, we used optical flow and good features as primary motion descriptors for the dynamic real time video obtained by camera attached to the SBC. The use of MLP network based on feed forward artificial neural network was done to train the system using KTH action dataset achieving an accuracy of 92%.

REFERENCES

[1] C. Schuldt, I. Laptev and B. Caputo, "Recognizing human actions: a local SVM approach," In *Proc. IEEE International Conference on Pattern Recognition (ICPR)*, 2004, pp. 32-36.
[2] I. Laptev and T. Lindeberg, "Space-time interest points," In *Proc. IEEE International Conference on Computer Vision (ICCV)*, Nice, France, 2003, pp. 432-439.
[3] G. Chéron, I. Laptev and C. Schmid, "P-CNN: Pose-Based CNN Features for Action Recognition," In *Proc. IEEE International Conference on Computer Vision (ICCV)*, Santiago, 2015, pp. 3218-3226.
[4] W. Yang, Y. Wang and G. Mori, "Human action recognition from a single clip per action," In *Proc. IEEE International Conference on Computer Vision Workshops*, Kyoto, 2009, pp. 482-489.
[5] Jianbo Shi and C. Tomasi, "Good features to track," In *Proc. IEEE Conference on Computer Vision and Pattern Recognition (ICVPR)*, Seattle, WA, 1994, pp. 593-600.
[6] K.Maithili, K. Rajeswari, R. Mohanapriya, D. Krithika, "An Efficient Human Action Recognition System Using Single Camera and Feature Points," International Journal of Advanced Research in Computer and Communication Engineering, Vol. 2, No. 2, 2013.
[7] R. Poppe, "A survey on vision-based human action recognition," Image and Vision Computing, Vol. 28, No. 6, pp. 976–990, 2010.
[8] A. F. Bobick and J. W. Davis, "The recognition of human movement using temporal templates," In IEEE Transactions on Pattern Analysis and Machine Intelligence, Vol. 23, No. 3, pp. 257-267, 2001.
[9] I. Laptev and T. Lindeberg, "Local descriptors for spatiotemporal recognition" In *Proc. ECCV Workshop on Spatial Coherence for Visual Motion Analysis*, 2004, pp. 91-103.
[10] J. C. Niebles, H. Wang, and L. Fei-Fei, "Unsupervised learning of human action categories using spatial-temporal words," International Journal of Computer Vision, Vol. 79, No. 3, pp. 299-318, 2006.
[11] T. Hofmann, "Unsupervised Learning by Probabilistic Latent Semantic Analysis," Machine Learning, Vol. 41, No. 2, pp. 177-196, 2001.
[12] D. Blei, A. Ng, and M. Jordan, "Latent Dirichlet allocation," Journal of Machine Learning Research, Vol. 3, pp. 993–1022, 2003.
[13] Li, Z. Stan, L. Zhu, Z. Zhang, A. Blake, H. Zhang, and H. Shum, "Statistical learning of multi-view face detection," In *Proc. European Conference on Computer Vision*, 2002, pp. 67-81.
[14] J. Fan, W. Xu, Y. Wu and Y. Gong, "Human Tracking Using Convolutional Neural Networks," In IEEE Transactions on Neural Networks, Vol. 21, No. 10, pp. 1610-1623, 2010.
[15] E. P. Ijjina and C. K. Mohan, "Human action recognition based on motion capture information using fuzzy convolution neural networks," In *Proc. International Conference on Advances in Pattern Recognition (ICAPR)*, Kolkata, 2015, pp. 1-6.
[16] P. KaewTraKulPong and R. Bowden, "An Improved Adaptive Background Mixture Model for Real-Time Tracking with Shadow Detection," In *Proc. European Workshop Advanced Video Based Surveillance Systems*, 2001.
[17] C. Harris and M. Stephens, "A Combined Corner and Edge Detector," In *Proc. Alvey Vision Conference*, Vol. 15. No. 50, pp. 147-151, 1988.
[18] J.Y. Bouguet, "Pyramidal Implementation of the Lucas Kanade Feature Tracker Description of the Algorithm," Intel Corporation, Vol. 5, No. 1-10, 2001.
[19] K. Guo, "Action recognition using log-covariance matrices of silhouette and optical-flow features," Ph.D. dissertation, Department of Electrical and Computer Engineering, Boston University, Boston, MA, Sep. 2011.
[20] Y. LeCun, L. Bottou, G.B. Orr, and K. Muller, "Efficient Backprop," Neural Networks, Tricks of the Trade, Springer, 1998.
[21] M. Riedmiller and H. Braun, "A direct adaptive method for faster backpropagation learning: the RPROP algorithm," *IEEE International Conference on Neural Networks*, San Francisco, CA, 1993, pp. 586-591.